\pdfoutput=1
\documentclass[11pt]{article}

\usepackage{EMNLP2023}

\usepackage{times}
\usepackage{latexsym}
\usepackage{graphicx}%
\usepackage{array}
\usepackage{multirow}%
\usepackage{amsmath,amssymb,amsfonts}%
\usepackage{amsthm}%
\usepackage{mathrsfs}%
\usepackage[title]{appendix}%
\usepackage{xcolor}%
\usepackage{textcomp}%
\usepackage{manyfoot}%
\usepackage{booktabs}%
\usepackage{hhline}
\usepackage{algorithm}%
\usepackage{algorithmicx}%
\usepackage{algpseudocode}%
\usepackage{listings}%
\usepackage[caption=false,font=normalsize,labelfont=sf,textfont=sf]{subfig}
\usepackage{stfloats}
\usepackage{url}
\usepackage{verbatim}
 \usepackage{booktabs}
 \usepackage{cite}
\usepackage{hyperref}
\hypersetup{
    colorlinks=true,
    linkcolor=black,
    filecolor=black,      
    urlcolor=black,
    citecolor=blue,
}
\usepackage{cleveref}
\usepackage{balance}
\usepackage{subcaption}
\newcounter{myverb}
\BeforeBeginEnvironment{verbatim}{%
\refstepcounter{myverb}%
\noindent\textbf{Prompt \themyverb}%
}

\usepackage[T1]{fontenc}
\usepackage[utf8]{inputenc}
\usepackage{microtype}
\usepackage{inconsolata}
\usepackage{newunicodechar}
\usepackage{placeins}
\newunicodechar{−}{-} % replace the actual U+2212 with ASCII hyphen

\title{BitMar: Low-Bit Multimodal Fusion with Episodic Memory for Edge Devices}

\author{
  Euhid Aman \\
  NTUST Taiwan \\
  \small \texttt{M11315803@mail.ntust.edu.tw}
  \And
  Esteban Carlin \\
  NTUST Taiwan \\
  \small \texttt{M11302809@mail.ntust.edu.tw}
  \And
  Hsing-Kuo Pao \\
  NTUST Taiwan \\
  \small \texttt{pao@mail.ntust.edu.tw}
  \AND
  Giovanni Beltrame \\
  Polytechnique Montréal \\
  \small \texttt{giovanni.beltrame@polymtl.ca}
  \And
  Ghaluh Indah Permata Sari \\
  NTUST Taiwan \\
  \small \texttt{d11115804@mail.ntust.edu.tw}
  \And
  Yie-Tarng Chen \\
  NTUST Taiwan \\
  \small \texttt{ytchen@mail.ntust.edu.tw}
}

\begin{document}
\maketitle

\begin{abstract}
Cross-attention transformers and other multimodal vision-language models excel at grounding and generation; however, their extensive, full-precision backbones make it challenging to deploy them on edge devices. Memory-augmented architectures enhance the utilization of past context; however, most works rarely pair them with aggressive edge-oriented quantization. We introduce BitMar, a quantized multimodal transformer that proposes an external human-like episodic memory for effective image-text generation on hardware with limited resources. BitMar utilizes 1.58-bit encoders, one for text (BitNet-style) and one for vision (DiNOv2-based), to create compact embeddings that are combined and used to query a fixed-size key-value episodic memory. During vector retrieval, the BitNet decoder applies per‑layer conditioning, which increases the contextual relevance of generated content. The decoder also employs attention sinks with a sliding‑window mechanism to process long or streaming inputs under tight memory budgets. The combination of per-layer conditioning and sliding-window attention achieves a strong quality–speed trade–off, delivering competitive captioning and multimodal understanding at low latency with a small model footprint. These characteristics make BitMar well-suited for edge deployment.

\end{abstract}

\section{Introduction}
\label{sec:intro}
Visual Language Models (VLMs) have made rapid progress in recent years, excelling at tasks such as image captioning~\citep{chen-microsoftcoco-2015}, visual question answering~\citep{anderson-attentionimagecap-2018, lijunan-blip-2022}. %and grounded dialogue generation~\citep{alayrac-flamingo-2022}. 
Large-scale architectures such as BLIP-2~\citep{lijunnan-blip2-2023}, Flamingo~\citep{alayrac-flamingo-2022}, and Kosmos-2~\citep{peng-kosmos2-2023} demonstrate that cross-attention transformers can synchronize modalities for grounded language generation. However, their full-precision, extensive backbones incur significant computational and memory expenses, which restricts their implementation on devices with resource limitations.

A growing body of work targets efficient multimodal processing, such as low-bit quantization~\citep{dettmers-8bitquantization-2021, frantar-gptq-2022} and compact language models~\citep{wang-bitnet-2023}, to reduce memory/latency. Quantized ViTs~\citep{jacob-quantization-2018, stock-revisiting-2019}, and self-supervised vision encoders, such as DiNOv2~\citep{oquab-dinov2-2024}, lower the cost of vision. Multimodal fusion ranges from early concatenation~\citep{lu-vilbert-2019} to learned query transformers~\citep{lijunnan-blip2-2023} to bridge frozen vision and language models. Memory-augmented transformers~\citep{graves-dynamicexternalmemory-2016, borgeaud-trillionstokens-2022} retrieve past context to improve coherence. 
Yet no existing tiny language model effectively unifies low-bit multimodal encoding with an episodic memory system for edge deployment.

To fill this gap, we propose a compact four-stage pipeline optimized for efficient on-device execution: 
(1) \textbf{1.58-bit text and vision encoders} generate lightweight, quantized embeddings; 
(2) \textbf{a cross-modal fusion module} aligns the modalities within a shared latent space; 
(3) \textbf{an episodic memory} with 512 key--value slots retrieves relevant multimodal context; and 
(4) \textbf{a BitNet-based decoder} conditions each transformer layer on the retrieved memory for context-aware generation. 
Both encoders output 128-dimensional representations, and DiNOv2’s original 768-D vision features are compressed to 128-D before fusion. 
The fused embedding queries an episodic memory of size $K = 512$, $C = 128$, whose retrieved vectors condition each decoder layer. 
This architecture maintains all modules in a consistent 768-dimensional space, simplifying integration and minimizing projection overhead while ensuring low-latency, memory-efficient operation on edge hardware.

Our main contributions are summarized as follows:
\begin{itemize}
    \item \textbf{Low-bit multimodal encoding framework.} 
    We propose a unified architecture that integrates a 1.58-bit quantized BitNet text encoder with a quantized ViT-based vision encoder, enabling efficient and compact multimodal feature extraction.
    
    \item \textbf{Memory-augmented decoding mechanism.} 
    We design a lightweight episodic memory module that retrieves contextual representations and injects them into each transformer layer through per-layer conditioning, enhancing coherence and contextual relevance during generation.
    
    \item \textbf{Edge-efficient multimodal reasoning.} 
    We demonstrate that BitMar achieves competitive performance in image captioning and multimodal understanding under extreme compression, maintaining low latency and a minimal memory footprint suitable for on-device deployment.
\end{itemize}

\section{Related Work}
\label{sec:relworks}

Different VLMs and Tiny LLM architectures have emerged that enable deployment and applications of multimodal AI on resource-constrained devices. Recent developments in small VLMs, such as H2OVL-Mississippi (0.8B parameters)~\citep{galib-2024-h2ovlmississippivisionlanguagemodels}, TinyGPT-V~\citep{tinygpt-v-2024}, and MiniCPM-V~\citep{minicpm-v-2024}, demonstrate that compact multimodal models can achieve competitive performance while maintaining efficient deployment characteristics. Similarly, Tiny LLMs, such as MobileLLM~\citep{mobilellm-2024} and TinyLLM~\citep{tinyllm-edge-2024}, have shown that sub-billion parameter models can be quantized and optimized for their deployment on edge devices. These highlight the feasibility of on-device multimodal processing, with models providing meaningful performance while addressing security, latency, and connectivity constraints.

Furthermore, memory-augmented neural networks and language models inspired by cognitive thinking, such as humans, have also garnered significant attention for their ability to store and retrieve contextual information related to specific things across certain short periods of time. 
Memory-augmented neural networks (MANNs)~\citep{graves-dynamicexternalmemory-2016}, use decoupled key-value structures to store and retrieve contextual information. Recent works, such as EGO~\citep{ego-episodic-2024} and selective episodic memory strategies~\citep{episodic-memory-neural-2022}, have extended these ideas for flexible knowledge transfer and context-based memory access. However, these models face limitations in combining memory systems with low-bit quantized multimodal encoders, often sacrificing either memory capacity or model precision.
% And recent works on similar episodic memory includes the Episodic Generalization and Optimization (EGO) framework, which integrates the episodic memory module, with semantic pathways to enable rapid learning and flexible knowledge transfer across tasks~\citep{ego-episodic-2024}. And Memory-Augmented Neural Networks (MANNs)~\citep{graves-dynamicexternalmemory-2016} have demonstrated improved performance through decoupled key-value structures that preserve explainability of historical knowledge extraction~\citep{decoupled-memory-2024}, while research on selective episodic memory encoding and retrieval shows that neural networks can learn human-like strategies for determining when to store and access memories based on environmental uncertainty and contextual relevance~\citep{episodic-memory-neural-2022}.

BitMar overcomes these challenges by integrating 1.58-bit quantization across text and vision encoders, alongside a cross-modal memory retrieval system. The design enables BitMar to store and retrieve both textual and visual context, improving memory interactions and enhancing multimodal generation tasks, all while maintaining computational efficiency for edge deployment.

\section{Method}
\label{sec:method}

\begin{figure*}[t]
    \centering
    \includegraphics[width=\textwidth]{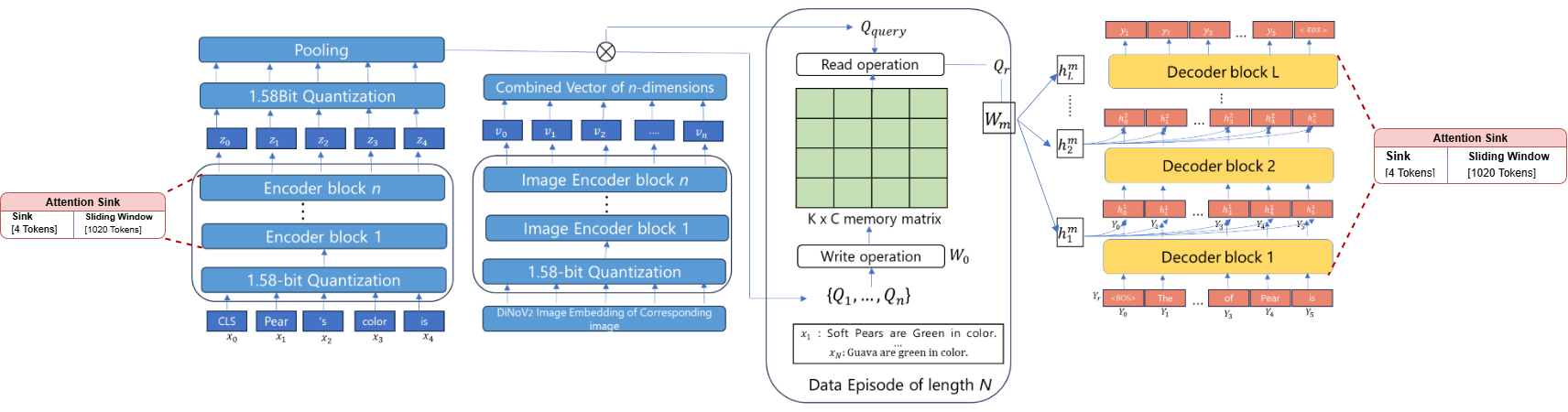}
    \caption{\textbf{BitMar Architecture.}
    The model processes multimodal inputs: text tokens and DiNOv2-compressed image features.
    Quantized encoders (1.58-bit) generate compact text and vision embeddings ($z$, $v$), which are fused via cross-modal attention into shared query representations ($Q$, $Q_{query}$).
    A sliding-window attention mechanism enables long-context processing.
    A fixed episodic memory matrix ($K \times C$) stores and retrieves multimodal context vectors through quantized read/write weights ($W$, $W_0$), supporting optional SD-card offloading for edge deployment.
    }
    % \caption{BitMar Architecture. $X$/$X_{query}$: multimodal inputs (text tokens + DiNOv2-compressed image features). Quantized encoders produce $z$/$v$ (1.58-bit). Cross-modal attention pools/fuses to $Q$/$Q_{query}$. Attention sinks with a sliding window (4 sink tokens, 1020 window) enable infinite-context processing. A fixed $K\times C$ episodic memory stores/retrieves episode vectors with quantized read/write weights $(W,W_0)$, supporting SD-card offloading.}
    \label{fig:bitmar}
\end{figure*}

We introduce \textbf{BitMar}, a deployable quantized multimodal LM for efficient image–text generation under tight resources. The four-stage pipeline is: 
(1) \textbf{parallel low-bit text/vision encoders}; 
(2) \textbf{cross-modal fusion} in a shared latent space; 
(3) \textbf{context augmentation} via external episodic memory; 
(4) \textbf{autoregressive decoding} conditioned on fused and retrieved signals. 
Text uses a BitNet transformer at 1.58-bit precision; vision uses DiNOv2 features plus quantization-aware compression. Fusion aligns 768-D modality latents via lightweight attention. A fixed-size episodic memory stores prior multimodal contexts and injects retrieved vectors into the decoder per layer. Unlike classic MANNs~\citep{graves-dynamicexternalmemory-2016}, BitMar integrates cross-modal retrieval under low-bit constraints. The decoder is a BitNet-based autoregressive transformer with streaming attention via attention sinks for low-latency, long-context generation.

% We present \textbf{BitMar}, a deployable quantized multimodal language model designed for efficient image–text generation under strict resource constraints. 
% The proposed architecture follows a four-stage pipeline: 
% (1) \textbf{parallel low-bit text and vision encoders} for compact representation learning; 
% (2) \textbf{cross-modal fusion} within a shared latent space to align textual and visual features; 
% (3) \textbf{context augmentation} through an external episodic memory that stores and retrieves multimodal context vectors; and 
% (4) \textbf{autoregressive decoding} conditioned on both fused embeddings and retrieved memory signals. 
% The text encoder adopts a 1.58-bit BitNet transformer, while the vision branch employs DiNOv2 features with quantization-aware compression. 
% Fusion aligns the 768-dimensional modality representations via lightweight attention. 
% A fixed-size episodic memory maintains previous multimodal contexts and injects the retrieved vectors into each decoder layer. 
% Unlike conventional memory-augmented neural networks~\citep{graves-dynamicexternalmemory-2016}, BitMar performs cross-modal retrieval under low-bit constraints, enabling a BitNet-based autoregressive decoder with streaming attention via attention sinks for low-latency, long-context generation.

\subsection{Text Encoders}

\textbf{Architecture.} A 4-layer quantized Transformer ($d{=}128$, $h{=}4$) supports up to 256 tokens, balancing expressiveness and efficiency.

\textbf{Quantization.} 
\emph{Weights:} all MHSA/FFN projections use ternary $\{-1,0,+1\}$ with learned per-layer scales (1.58-bit). 
\emph{Activations:} token-wise 8-bit using per-token max-abs scaling to $[-127,127]$, preserving local detail and stable training/inference.

\textbf{Attention sinks (streaming).} With $S{=}4$ sink tokens (never evicted) and window $W{=}1020$, the KV cache maintains persistent anchors + recent tokens. On each new token, the oldest in-window token is evicted; sink and window sets are merged; positions are clamped to $[0,S{+}W{-}1]$. This yields fixed-memory, long-context attention under low-bit compute.

\subsection{Vision Encoders}
We use frozen DiNOv2~\citep{oquab-dinov2-2024} to extract 768-D patch features offline, avoiding heavy vision backbones at inference. $2{\times}2$ average pooling reduces the number of patches $4{\times}$ while keeping 768-D per patch. 2-layer MLP bottleneck then compresses 768$\rightarrow$128 with ReLU and dropout between layers (parameters $\mathbf{W_1}\in\mathbb{R}^{384\times 768}, \mathbf{W_2}\in\mathbb{R}^{128\times 384}$), all subsequent fusion/memory/decoder paths operate in 128-D.

% \textbf{Spatial pooling.} $2{\times}2$ average pooling reduces patches $4{\times}$ while keeping 768-D per patch.

% \textbf{Compression.} A 2-layer MLP bottleneck maps 768$\rightarrow$128:
% \begin{align}
%     \mathbf{z}^{(1)}_{\text{vis}} &= \mathrm{ReLU}(\mathbf{W}_1 \mathbf{x}_{\text{vis}} + \mathbf{b}_1), \\
%     \mathbf{z}^{(2)}_{\text{vis}} &= \mathbf{W}_2 \mathbf{z}^{(1)}_{\text{vis}} + \mathbf{b}_2
% \end{align}
% with dropout after ReLU.

\subsection{Cross-Modal Fusion}
Given pooled text tokens $\mathbf{Z}\in\mathbb{R}^{n_t\times128}$ and vision tokens
$\mathbf{V}_{\text{img}}\in\mathbb{R}^{n_v\times128}$, we apply standard cross-attention~\citep{vaswani2017attention}
(text queries, vision keys/values; cf.\ Transformer attention) to obtain the fused
sequence $\mathbf{F}\in\mathbb{R}^{n_t\times128}$. All $Q/K/V$ and fusion projections
use 1.58-bit ternary weights with learned scales; softmax and residual/LN are in FP32.
We then pool $\mathbf{F}$ (mean or learned) to a single vector
$\mathbf{q}_{\text{mem}}\in\mathbb{R}^{128}$ to query episodic memory.

% Given pooled outputs $\mathbf{Z}\in\mathbb{R}^{n\times128}$ (text) and $\mathbf{V}\in\mathbb{R}^{n\times128}$ (vision), we apply cross-attention with text queries and vision keys/values:
% \begin{equation}
%     \mathrm{Attention}(Q,K,V)=\mathrm{softmax}\!\left(\frac{QK^\top}{\sqrt{d_k}}\right)V
% \end{equation}
% All projections for $Q,K,V$ and the fused output use 1.58-bit ternary weights with learned scales; softmax and residual/LN run in FP32. The fused sequence $\mathbf{F}\in\mathbb{R}^{n\times128}$ is pooled (mean or learned) to form $\mathbf{Q_{query}}\in\mathbb{R}^{n\times128}$ for memory retrieval.

\subsection{Episodic Memory}

We maintain a learnable matrix $\mathbf{M}\in\mathbb{R}^{K\times C}$ (default $K{=}512$, $C{=}128$) that stores multimodal episode vectors.

\paragraph{Writing.}
At step $t$, we compute a pooled query $\mathbf{q}_t\in\mathbb{R}^{C}$ and learned write weights $\mathbf{W}_w\in\mathbb{R}^{K}$. We perform soft multi-slot writes with rate $\alpha{=}0.2$ via an outer product:
\begin{equation}
\mathbf{M} \leftarrow \mathbf{M} + \alpha\, \mathbf{W}_w \,\mathbf{q}_t^{\top}.
\label{eq:mem_write}
\end{equation}

\paragraph{Reading.}
We use content-based addressing \citep{graves-dynamicexternalmemory-2016}:
\begin{equation}
\begin{aligned}
\mathbf{W}_r &= \operatorname{softmax}(\mathbf{M}\,\mathbf{q}_t)\in\mathbb{R}^{K},\\
\mathbf{M}_r &= \mathbf{W}_r^{\top}\mathbf{M}\in\mathbb{R}^{1\times C}.
\end{aligned}
\label{eq:mem_read}
\end{equation}

\paragraph{Regularization.}
To avoid thrashing, we penalize abrupt updates to the store with a Frobenius penalty,
% \begin{equation}
$
\mathcal{L}_{\mathrm{reg}} = \lambda \,\big\lVert \Delta\mathbf{M}\big\rVert_{F}^{2},
\quad
\Delta\mathbf{M} := \mathbf{M}^{(t)} - \mathbf{M}^{(t-1)}.
\label{eq:mem_reg}
$
% \end{equation}
We additionally apply usage-based forgetting to down-weight stale slots.
% A learnable matrix $\mathbf{M}\in\mathbb{R}^{K\times C}$ ($K{=}512$, $C{=}128$) stores multimodal episode vectors; 512 slots worked best empirically.

% \textbf{Writing.} For step $t\in[[1,n]]$, write with learned weights $\mathbf{W_w}\in\mathbb{R}^K$:
% \begin{equation}
% \mathbf{M} \leftarrow \mathbf{M} + \alpha \cdot \mathbf{W}w^\top \mathbf{Q}_{\text{query},t}
% \end{equation}
% with update rate $\alpha{=}0.2$ (soft multi-slot writes).

% \textbf{Reading.} Content addressing:
% \begin{equation}
% \mathbf{W}r = \mathrm{softmax}(\mathbf{M}\mathbf{Q}_{\text{query}})
% \end{equation}
% and retrieved vector
% \begin{equation}
% \mathbf{M}_r = \mathbf{W}_r^\top \mathbf{M}\in\mathbb{R}^{1\times C}.
% \end{equation}

% \textbf{External storage.} $\mathbf{M}$ can be offloaded (e.g., external SD card) to persist state without GPU memory growth.

% \textbf{Regularization.} We penalize updates with
% \[
% \mathcal{L}_{\text{reg}}=\lambda\|\Delta\mathbf{M}\|_2^2
% \]
% and apply usage-based forgetting to down-weight stale slots.

\subsubsection{Decoder with Attention Sinks}
A 4-layer causal Transformer ($d{=}128$, $h{=}4$, max length 256) conditions on fused inputs and retrieved memory.

\textbf{Long-context generation.} Each layer, similarly as the text encoder, maintains KV caches of $S$ sink tokens and a window of $W$ recent tokens.

\textbf{Memory integration.} $\mathbf{M_r}\in\mathbb{R}^{1\times128}$ is projected and combined with token embeddings via either concatenation $[x_t;\mathbf{M_r}]$ (then projected) or residual addition $x_t{+}\mathbf{M_r}$.

\textbf{Output projection.} BitNet-quantized linear layer (128$\rightarrow$50{,}257) maps to GPT-2 vocab logits; logits computed in FP32.

\subsubsection{Training Objectives}
We complement standard Language Modeling cross-entropy~\citep{vaswani2017attention} and an InfoNCE cross-modal~\citep{oord_representation_2018} term with a memory-consistency regularizer~\autoref{eq:memory_consistent} that penalizes changes between successive writes to the episodic store, which discourages oscillatory updates and helps retain slot semantics. The total loss integrates these factors as~\autoref{eq:total_obj}. We set $\mathcal{L}_{\text{cm}}=1.5$ to prioritize cross-modal alignment, and $\mathcal{L}_{\text{mem}}=0.1$ as a light stabilizer.
% \textbf{Language modeling.} We optimize a composite objective with three terms. First, we use standard token-level cross-entropy for language modeling~\citep{vaswani2017attention}. Second, we align pooled text/vision embeddings with an InfoNCE cross-modal contrastive objective~\citep{oord_representation_2018}. Third, we complement a memory-consistency penalty~\autoref{eq:memory_consistent}. Lastly, we compute the total loss as a weighted sum~\autoref{eq:total_obj}:
% \begin{equation}
% \mathcal{L}_{\text{lm}} = - \frac{1}{T} \sum_{t=1}^T \log p_{\theta}(y_t^\ast \mid y_{<t}, \mathbf{X})
% \end{equation}

% \textbf{Cross-modal contrastive (InfoNCE)~\citep{oord_representation_2018}}
% \small
% \begin{equation}
% \mathcal{L}_{\text{cm}} = - \frac{1}{N} \sum_{i=1}^N \log \frac{\exp(\mathrm{sim}(\mathit{z}_t^i, \mathit{z}_v^i) / \tau)}{\sum_{j=1}^N \exp(\mathrm{sim}(\mathit{z}_t^i, \mathit{z}_v^j) / \tau)}
% \end{equation}
% \normalsize

\textbf{Memory consistency.}
\begin{equation}
\mathcal{L}_{\text{mem}} = | \mathbf{M}^{(t)}_{\text{write}} - \mathbf{M}^{(t-1)}_{\text{write}} |_2^2
\label{eq:memory_consistent}
\end{equation}

\textbf{Total objective.}
\begin{equation}
\mathcal{L} = \mathcal{L}_{\text{lm}} + 1.5 · \mathcal{L}_{\text{cm}} + 0.1 · \mathcal{L}_{\text{mem}}
\label{eq:total_obj}
\end{equation}

% \subsubsection{Adaptive Training Controller}
% We stabilize alignment via an \textbf{Adaptive Training Controller (ATC)} that monitors the exponential moving average (EMA) of cross-modal cosine similarity over a 200-step window.

% \textbf{Trigger.} Intervene on a relative EMA drop $>0.12$, from the recent maximum and, if $\geq$800 steps since the last intervention.

% \textbf{Actions.} Randomly: freeze text encoder; freeze vision encoder; or boost the weight of $\mathcal{L}_{\text{cm}}$. The chosen action persists for 1{,}500 steps.

% \textbf{Effect.} ATC mitigates modality collapse by enforcing balanced contributions to the shared space and maintaining retrieval/generation quality under challenging optimization.

\textbf{Adaptive Training Controller.} When a 200-step EMA of cross-modal cosine similarity drops by $>0.12$ from its recent max (with an $\geq$800-step cooldown), we randomly freeze one encoder or upweight $\mathcal{L}_{\text{cm}}$ for 1,500 steps to prevent modality collapse.

\section{Experimental Setup}
\label{sec:exp}

Our experimental framework systematically evaluates the proposed 14M-parameter BitMar model across several critical dimensions. We first benchmark its performance against established compact and low-bit baselines to assess overall viability (\autoref{tab:ranking_metrics}). We then conduct an analysis of its capabilities across a suite of language understanding and multimodal tasks to identify specific strengths and limitations (\autoref{tab:bitmar_babylm_results}). Beyond task performance, we also investigate the internal dynamics of the model, examining how the episodic memory evolves from diffuse to structured activation patterns during training (\autoref{fig:heatmap_comparison}). Finally, we track the progression of quantization efficacy throughout the training process to validate our low-precision approach (\autoref{fig:quantization}).

\subsection{Dataset}
The corpus comprises 100M tokens, split evenly between multimodal captions and text-only data.  

\textbf{Multimodal (50M).} From CC3M~\citep{sharma-CC3M-2018} and Localized Narratives~\citep{ponttuset-localized-narratives-2020}, aligned with precomputed DiNOv2 features (frozen backbone, reused across training).  

\textbf{Text-only (50M).} From BabyLM~\citep{Charpentier-babylm-2025}, spanning six domains (BNC, CHILDES, Gutenberg, OpenSubtitles, Simple English Wikipedia, Switchboard).  

\textbf{Mixture.} Uniform 50:50 sampling; a 1M-token hold-out tracks cross-modal alignment (cosine similarity) and perplexity.  

\textbf{Preprocessing.} GPT-2 BPE tokenizer, max 256 tokens (truncate/pad). Visual features stored as memory-mapped ``.npy'' with on-the-fly compression for efficient batching.

\subsection{Training Configuration}
We trained on an NVIDIA A6000 GPU using FP16 and gradient checkpointing. Each step processed 64 sequences, with two-step gradient accumulation yielding an effective batch size of 128. Optimization used AdamW8bit ($2\!\times\!10^{-4}$) with cosine restarts ($T_0{=}1000$, $T_{\text{mult}}{=}2$, $\eta_{min}{=}0.1lr$) for 10 epochs. We logged to Weights \& Biases every 500 steps, including losses ($\mathcal{L}_{\text{lm}}, \mathcal{L}_{\text{cm}}, \mathcal{L}_{\text{mem}}$), cross-modal alignment metrics, episodic-memory utilization, attention maps, and FLOPs per step.
% Experiments were run on an A5000 GPU using FP16 and gradient checkpointing.  

% \textbf{Batching.} 64 sequences per step, adequate batch size 128 via 2-step accumulation.  

% \textbf{Optimization.} AdamW8bit ($2\!\times\!10^{-4}$) with cosine restarts ($T_0{=}1000$, $T_{\text{mult}}{=}2$, $\eta_{min}{=}0.1lr$) for 10 epochs.  

% \textbf{Monitoring.} Weights \& Biases logging every 500 steps: losses ($\mathcal{L}_{\text{lm}}, \mathcal{L}_{\text{cm}}, \mathcal{L}_{\text{mem}}$), alignment metrics, episodic usage, attention maps, and FLOPs/step.

\subsection{Hyperparameters}
% \autoref{tab:hyperparams} lists key settings.
The model architecture employs a four-layer text encoder with 128-dimensional hidden states. The episodic memory module comprises 512 slots, each with 128 dimensions, balancing memory footprint with recall capacity. For long-context streaming, we maintain four sink tokens with a sliding window of 1020 tokens. Training utilizes weighted losses with cross-modal and memory consistency coefficients of 1.5 and 0.1, respectively. An adaptive controller triggers memory freezing when alignment metrics drop by 0.12 from their recent maximum, applying 1,500-step freezes with a minimum interval of 800 steps between interventions.

% \begin{table}[ht]
% \centering
% \small
% \begin{tabular}{l l}
% \hline
% \textbf{Hyperparameter} & \textbf{Value} \\
% \hline
% Text Encoder layers         & 4 \\
% Text hidden dim             & 128 \\
% Memory slots                & 512 \\
% Memory dim                  & 128 \\
% Sink size                   & 4 \\
% Window size                 & 1,020 \\
% Cross-modal loss weight     & 1.5 \\
% Memory consistency weight   & 0.1 \\
% Drop threshold (adaptive)   & 0.12 \\
% Freeze duration (steps)     & 1,500 \\
% Min steps per intervention  & 800 \\
% \hline
% \end{tabular}
% \caption{Training hyperparameters.}
% \label{tab:hyperparams}
% \end{table}

\subsection{Benchmarks and Baselines}
We evaluate on six language benchmarks: \textit{ARC-Easy}, \textit{BoolQ}, \textit{HellaSwag}, \textit{WinoGrande}, \textit{CommonsenseQA}, and \textit{MMLU}, plus multimodal tasks aligned with DiNOv2 features. Outputs are evaluated by accuracy and compared against baselines (\textit{Bonsai 0.5B}, \textit{OLMo-BitNet 1B}, \textit{Falcon3-1.58bit 7B}, \textit{LLaMA3-8B-1.58}, and \textit{BitNet b1.58 2B}). Beyond benchmarks, we track the effectiveness of quantization and episodic activations to assess representational efficiency and memory use.

\section{Results and Discussion}
\label{sec:result}
\subsection{BitMar's performance}

\begin{figure}[!htbp]
     \centering
     \begin{minipage}[t]{.45\linewidth}
    \includegraphics[width=\linewidth]{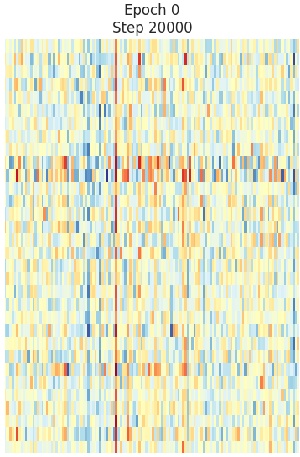}\\
        \centering(a)
    \end{minipage}\hfill%
    \begin{minipage}[t]{.45\linewidth}
    \includegraphics[width=\linewidth]{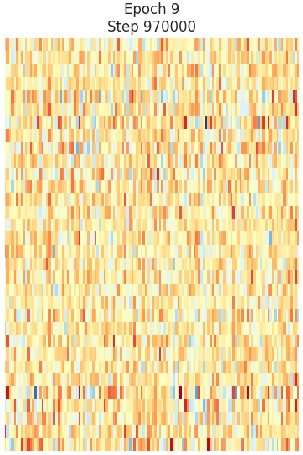}\\
        \centering(b)
    \end{minipage}
    \caption{\textbf{Episodic Memory Activation Patterns.}
    (a) Early training shows scattered and weak activations with minimal specialization.
    (b) Late training exhibits stronger and more differentiated activations, reflecting the emergence of structured memory representations.
    }
    \label{fig:heatmap_comparison}
\end{figure}

~\autoref{fig:heatmap_comparison} shows episodic memory slot activations over training. Early on~\autoref{fig:heatmap_comparison}(a), activations are weak and scattered, with minor specialization or proper storage. By late training~\autoref{fig:heatmap_comparison}(b), activations strengthen and differentiate, indicating selective storage of contextual features. This progression demonstrates that extended joint optimization enables the memory to evolve into a more structured, capacity-efficient component for long-term context integration.

We measure the quantization effectiveness $E_q$, inspired by~\citep{zhu-tenary_quantization-2016}, as the zero-weight fraction in ternary weights across BitNet-quantized layers, where a higher value means more compression. 

% $E_q$ increases to 42.8\% (\autoref{fig:quantization}), step-like early, then smooth, without degrading downstream performance.

\begin{figure}[!htbp] 
	\centering
	\includegraphics[width=\linewidth]{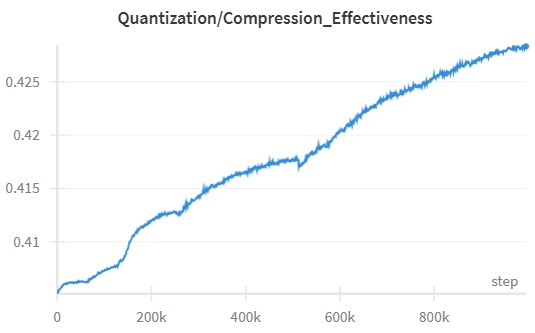}
	\caption{\textbf{Quantization effectiveness over training epochs.}
    } 
    %Compression effectiveness (fraction of quantized weights set to zero). A value of 42.8\% indicates efficient quantization with substantial representational preservation.
	\label{fig:quantization}
\end{figure}

As training progresses (\autoref{fig:quantization}), $E_q$ gradually increases and stabilizes at 42.8\%, demonstrating effective compression without degrading downstream performance.
% The quantization effectiveness metric (Figure~\ref{fig:quantization}) steadily rises to 42.8\%, reflecting representational quality relative to storage/computation cost. Initial improvements occur in discrete jumps (0–200k steps), likely from adaptation in text and vision encoders; after 400k steps, growth smooths, suggesting convergence. The final score shows that BitMar learns dense representations within its quantized space, retaining much of its multimodal capacity while greatly reducing memory and compute.

\begin{table*}[t]
\centering
\small
\begin{tabular}{lccccccc}
\hline
\textbf{Model} & \textbf{Native 1-bit} & \textbf{ARC-Easy} & \textbf{BoolQ} & \textbf{HellaSwag} & \textbf{WinoGrande} & \textbf{CommonsenseQA} & \textbf{MMLU} \\
\hline
Bonsai 0.5B        & \checkmark & 58.25 & 58.44 & 48.01 & 54.46 & 18.43 & 25.74 \\
OLMo-BitNet 1B     & \checkmark & 25.38 & 52.48 & 25.88 & 51.54 & 19.49 & 25.47 \\
Falcon3-1.58bit 7B & $\times$   & 65.03 & 72.14 & 59.46 & 60.14 & 67.08 & 42.79 \\
LLaMA3-8B-1.58 8B  & $\times$   & 70.71 & 68.38 & 68.56 & 60.93 & 28.50 & 35.04 \\
BitNet b1.58 2B    & \checkmark & 74.79 & 80.18 & 68.44 & 71.90 & 71.58 & 53.17 \\
BitMar-14M (Ours)  & \checkmark & 28.32 & 42.83 & 30.04 & 54.57 & 24.57 & 27.90 \\
\hline
\end{tabular}
\caption{\textbf{Benchmark performance on language understanding tasks.} 
A \checkmark indicates models trained natively with 1-bit precision. 
All reported values correspond to task accuracy (\%), illustrating BitMar’s competitive performance under extreme compression.
}
\label{tab:ranking_metrics}
\end{table*}

~\autoref{tab:ranking_metrics} compares BitMar-14M with low-bit baselines. Despite its small size (14M parameters), it achieves competitive performance on \textit{BoolQ} (42.8) and \textit{WinoGrande} (54.6), demonstrating strength in binary reasoning and coreference. On \textit{ARC-Easy} (28.3) and \textit{HellaSwag} (30.0), it lags larger models, reflecting limits in multi-step reasoning. \textit{CommonsenseQA} (24.6) and \textit{MMLU} (27.9) remain challenging due to restricted factual coverage. Still, BitMar achieves non-trivial accuracy across all tasks, confirming that extreme compression can yield usable models for targeted workloads, though with expected trade-offs in knowledge-heavy benchmarks.

\begin{table*}[t]
\centering
\small
\begin{tabular}{l l l l}
\hline
\textbf{Category} & \textbf{Task} & \textbf{Primary Metric} & \textbf{Score} \\
\hline
Finetune NLP & BoolQ & Accuracy & 66.5\% \\
             & MNLI & Accuracy & 42.3\% \\
             & MRPC & Accuracy & 69.1\% \\
             & MultiRC & Accuracy & 57.6\% \\
             & QQP & Accuracy & 70.2\% \\
             & RTE & Accuracy & 54.0\% \\
             & WSC & Accuracy & 63.5\% \\
Multimodal   & DevBench & Visual Vocab Acc. & 21.2\% \\
             & VQA & Accuracy & 21.4\% \\
             & Winoground & Accuracy & 23.8\% \\
World Knowledge & EWOK & Accuracy & 24.9\% \\
Linguistic   & BLIMP & Accuracy & 48.7\% \\
Reasoning    & Compositional & Accuracy & 51.5\% \\
             & Entity Tracking & Accuracy & 31.2\% \\
Psycholing.  & Reading Comp. & Score & 0.44 \\
Morphology   & Wug Adj. & Corr. & -0.16 \\
             & Wug Past & Corr. & -0.22 \\
\hline
\end{tabular}
\caption{BitMar results on BabyLM evaluation tasks.}
\label{tab:bitmar_babylm_results}
\end{table*}

As shown in~\autoref{tab:bitmar_babylm_results}, BitMar achieves an average 60.5\% across finetuned NLP benchmarks, with strong results on paraphrase (\textit{QQP}: 70.2\%, \textit{MRPC}: 69.1\%) and reading comprehension (\textit{BoolQ}: 66.5\%), but weaker performance on inference (\textit{MNLI}: 42.3\%, \textit{RTE}: 54.0\%). Multimodal tasks yield modest scores (21–25\%), with the best results on \textit{EWoK} (24.9\%), likely benefiting from episodic memory. Linguistic analysis shows reasonable syntax (\textit{BLIMP}: 48.7\%) and compositional reasoning (51.5\%), but poor morphological productivity (\textit{WUG}: −0.16/−0.22). Overall, BitMar balances extreme efficiency with usable performance, excelling in lightweight reasoning while struggling on complex multimodal and morphological tasks.

\subsection{Ablation Study: Episodic Memory}

Evaluated under BabyLM 2025 evaluation pipeline (same as \autoref{tab:bitmar_babylm_results}).

\paragraph{Efficiency.}
As~\autoref{tab:inference_ablation_metrics} reports, a fixed retrieved vector supplies context each step, reducing long-range attention while keeping 1.58-bit compute.

\begin{table}[t]
    \centering
    % \begin{subtable}{0.95\linewidth}
    \begin{tabular}{lcc}
        \hline
        \textbf{Metric} & \textbf{Mem. On} & \textbf{Mem. Off} \\ 
        \hline
        Throughput (tok/s) & 57.3 & 7.7 \\
        Latency/token (ms) & 17.3 & 129.8 \\
        Energy (J) & 1.90 & 9.17 \\
        RAM (MB) & 956 & 1,076 \\
        \hline
    \end{tabular}
    \caption{\textbf{Inference ablation metrics.}
    Comparison of throughput, latency, energy consumption, and memory usage.
    }
    \label{tab:inference_ablation_metrics}
\end{table}
\vspace{-4pt}
\begin{table}[t]
    \centering
     \begin{tabular}{lc}
        \hline
        \textbf{Task} & \textbf{$\Delta$ (pp)} \\ 
        \hline
        Entity Tracking (Split 1) & +2.9 \\
        Entity Tracking (Split 2) & +4.1 \\
        COMPS & +3.4 \\
        BLiMP & +0.6 \\
        VQA & +3.4 \\
        EWoK (Split 1) & $-$1.6 \\
        EWoK (Split 2) & +1.0 \\
        Winoground & $-$1.6 \\
        DevBench & No effect \\
        \hline
    \end{tabular}
    \caption{\textbf{Ablation results on episodic memory.}
    Performance differences ($\Delta$, in percentage points), positive values indicate improvements when memory is enabled.
    }
\label{tab:ablation_results}
\end{table}

\paragraph{Zero-shot accuracy in $\Delta$ (pp).}
\autoref{tab:ablation_results} reports the performance differences on zero-shot tasks. Overall, the results suggest that incorporating additional contextual information generally enhances task accuracy.

\paragraph{Regressions.}
We observe two regressions. First, regarding \textit{WUG} morphology, correlations are negative, $-$0.36 for adjectives and $-$0.16 for past tense, indicating reduced morphological productivity under extreme quantization. Second, \textit{reading} alignment scores are lower with memory (0.44/0.11) than without (1.11/0.66), suggesting that episodic conditioning can dampen psycholinguistic alignment. Tuning memory capacity or injection strategy may mitigate this.
% \begin{itemize}
%   \item \textbf{WUG morphology:} $-$0.36 (adj), $-$0.16 (past)
%   \item \textbf{Reading alignment:} lower with memory (0.44/0.11 vs.\ 1.11/0.66)
% \end{itemize}

\paragraph{Fine-tuning.}
No significant changes on \textit{BoolQ/MNLI/MRPC/MultiRC/QQP/RTE/WSC}, suggesting memory mainly affects generation, not supervised heads.

\paragraph{Ablation Summary.} Episodic Memory is $\sim$7.5$\times$ faster, using 79\% less energy and 11\% less VRAM in our tests. It delivers 3 – 4 percentage point gains on entity/property reasoning and multimodal QA, though morphology and some psycholinguistic alignment metrics can degrade. Overall, combining attention sinks with episodic memory enables efficient long-context use under tight resource budgets.
% \paragraph{Takeaways.}
% \begin{enumerate}
%   \item \textbf{Efficiency:} $\sim$7.5$\times$ faster, 79\% less energy, 11\% less RAM, supporting edge use.  
%   \item \textbf{Helps:} +3--4 pp on entity, property, multimodal QA.  
%   \item \textbf{Doesn't help:} Morphology and psycholinguistic alignment can degrade.  
%   \item \textbf{Overall:} Attention-sinks with Episodic Memory enables efficient context use.
% \end{enumerate}
\section{Conclusion}
\label{sec:conclusion}
\textbf{BitMar-14M} is a compact 1.58-bit multimodal language model using BitNet quantization, DiNOv2 vision compression, cross-modal fusion, an attention-sink decoder for efficient long-context reasoning, and an external episodic latent memory for deployment on resource-constrained edge devices. With adaptive training, it maintains stable alignment and memory use despite its tiny size. Though less accurate than larger low-bit models on knowledge-heavy tasks, it performs competitively on binary reasoning and coreference, showing that 1.58-bit compression and efficient design can enable multimodal reasoning with drastically reduced compute and storage.

\FloatBarrier

\bibliographystyle{acl_natbib}
\bibliography{bitmar}

% \bibliographystyle{acl_natbib}

%Please see Section~\ref{sec:bibtex} for information on preparing Bib\TeX{} files.

% \section{Bib\TeX{} Files}
% \label{sec:bibtex}

% \bibliographystyle{acl_natbib}

% \appendix

% \section{Example Appendix}
% \label{sec:appendix}

% This is a section in the appendix.

\end{document}